\pdfoutput=1

\documentclass[letterpaper, 10 pt, conference]{ieeeconf}  

\IEEEoverridecommandlockouts                              

\usepackage{graphicx}
\graphicspath{ {Figures/}  {Tables/} }
\usepackage{float}
\usepackage[binary-units=true]{siunitx}
\usepackage{cite}
\usepackage{amsmath} 
\usepackage{amssymb}  
\newcolumntype{R}[1]{>{\RaggedLeft\arraybackslash}p{#1}}
\usepackage{hyperref}

\usepackage[utf8]{inputenc}
\usepackage{amsmath}
 
\usepackage{flushend}

\begin{document}

\title{\LARGE \bf
A Fog Robotic System for Dynamic Visual Servoing}

\author{Nan Tian,$^{1,2}$ Jinfa Chen$^{2}$, Mas Ma$^{2}$, Robert Zhang$^{2}$, Bill Huang$^{2}$ \\
Ken Goldberg$^{1}$ and Somayeh Sojoudi$^{1, 3, 4}$}

\thanks{$^{1}$Department of Electrical Engineering and Computer Science (EECS), University of California, Berkeley, USA; \{{\tt\small neubotech, goldberg, sojoudi}\}{\tt\small@berkeley.edu}}
\thanks{$^{2}$Advanced Technology Department, Cloudminds Technology Inc., Santa Clara, US; \{{\tt\small robert, bill}\}{\tt\small@cloudminds.com}}
\thanks{$^{3}$Department of Mechanical Engineering, University of California, Berkeley, USA;}
\thanks{$^{4}$Tsinghua-Berkeley Shenzhen Institute.}

\maketitle

\begin{abstract}
Cloud Robotics is a paradigm where distributed robots are connected to cloud services via networks to access ``unlimited” computation power, at the cost of network communication.  However, due to limitations such as network latency and variability, it is difficult to control dynamic, human compliant service robots directly from the cloud.  In this work, by leveraging asynchronous protocol with a “heartbeat” signal, we combine cloud robotics with a smart edge device to build a Fog Robotic system.  We use the system to enable robust teleoperation of a dynamic self-balancing robot from the cloud.  We first use the system to pick up boxes from static locations, a task commonly performed in warehouse logistics.  To make cloud teleoperation more efficient, we deploy image based visual servoing (IBVS) to perform box pickups automatically.  Visual feedbacks, including apriltag recognition and tracking, are performed in the cloud to emulate a Fog Robotic object recognition system for IBVS.  We demonstrate the feasibility of real-time dynamic automation system using this cloud-edge hybrid, which opens up possibilities of deploying dynamic robotic control with deep-learning recognition systems in Fog Robotics.  Finally, we show that Fog Robotics enables the self-balancing service robot to pick up a box automatically from a person under unstructured environments.
\end{abstract}

\section{INTRODUCTION}

 Service robots are robots that operate semi- or fully autonomously to perform services useful to the well-being of humans and equipments\cite{IFRServiceRobot}.  International Federation of Robotics (IFR) predicts that 32 million service robots are to be deployed between 2018-2022 \cite{IFRServiceRobotMarket}. Some popular service robot applications include elderly care, house cleaning, cooking, patrol robots, robot receptionists, entertainment, and education. A few famous examples of service robots are Roomba by iRobot, Pepper by Softbank Robotics, “the robotic chef” by Moley Robotics, and Spotmini and Atlas by Boston Dynamics.
 
 Different from industrial robots, service robots need to interact and cooperate with people safely under dynamic unstructured environments.  Therefore, two key requirements for service robot operations are (1) accurate, general visual perceptions and (2) intelligent, dynamic, human compliant robot controls.
 
 \begin{figure}
 	\centering
 	\includegraphics[width=1.0\columnwidth] {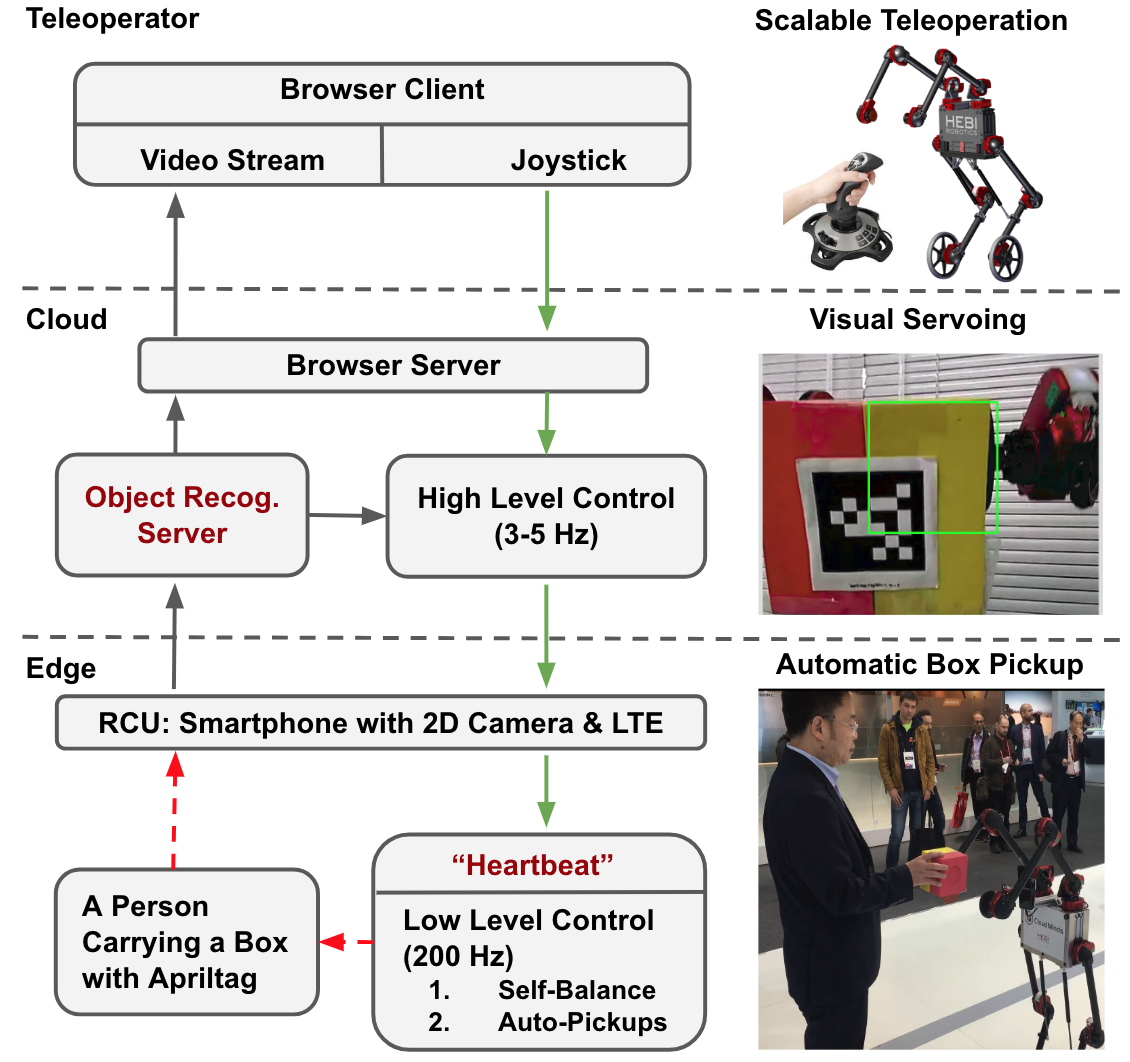}
 	\caption{ \textbf{The Fog Robotic system for dyanmic visual servoing:}
 		\textbf{(Left)} Architecture diagram and information flow--visual perceptions (black arrows), control signals (green arrows), human-robot interactions (red arrows).
 		\textbf{(Right)} Illustrations of three major contributions of this work, teleoperation, visual servoing, and auto box-pickups from a human.  They are positioned to their related functional blocks in the Fog Robotic system}
 	\label{fig:archetecture}
 \end{figure}

With recent breakthroughs in deep neural networks and robotic learning, robot visual perceptions \cite{girshick2015fast}\cite{krizhevsky2012imagenet}\cite{matterport_maskrcnn_2017}\cite{cao2017realtime} and intelligent controls \cite{mahler2017dex}\cite{levine2014learning}\cite{levine2016learning}\cite{finn2017deep}\cite{lee2017learning} under unstructured environments have become readily available .  However, these learning-based technologies come with a high computation cost, and it is hard to deploy them directly on native robot controllers that have limited computation power.  In our previous work on gesture based semaphore mirroring using a humanoid robot \cite{tian12cloud}, we addressed this problem by moving deep-learning-based gesture inferencing into the cloud.

\begin{figure*} [th]
	\centering
	\includegraphics[width=0.75\textwidth]{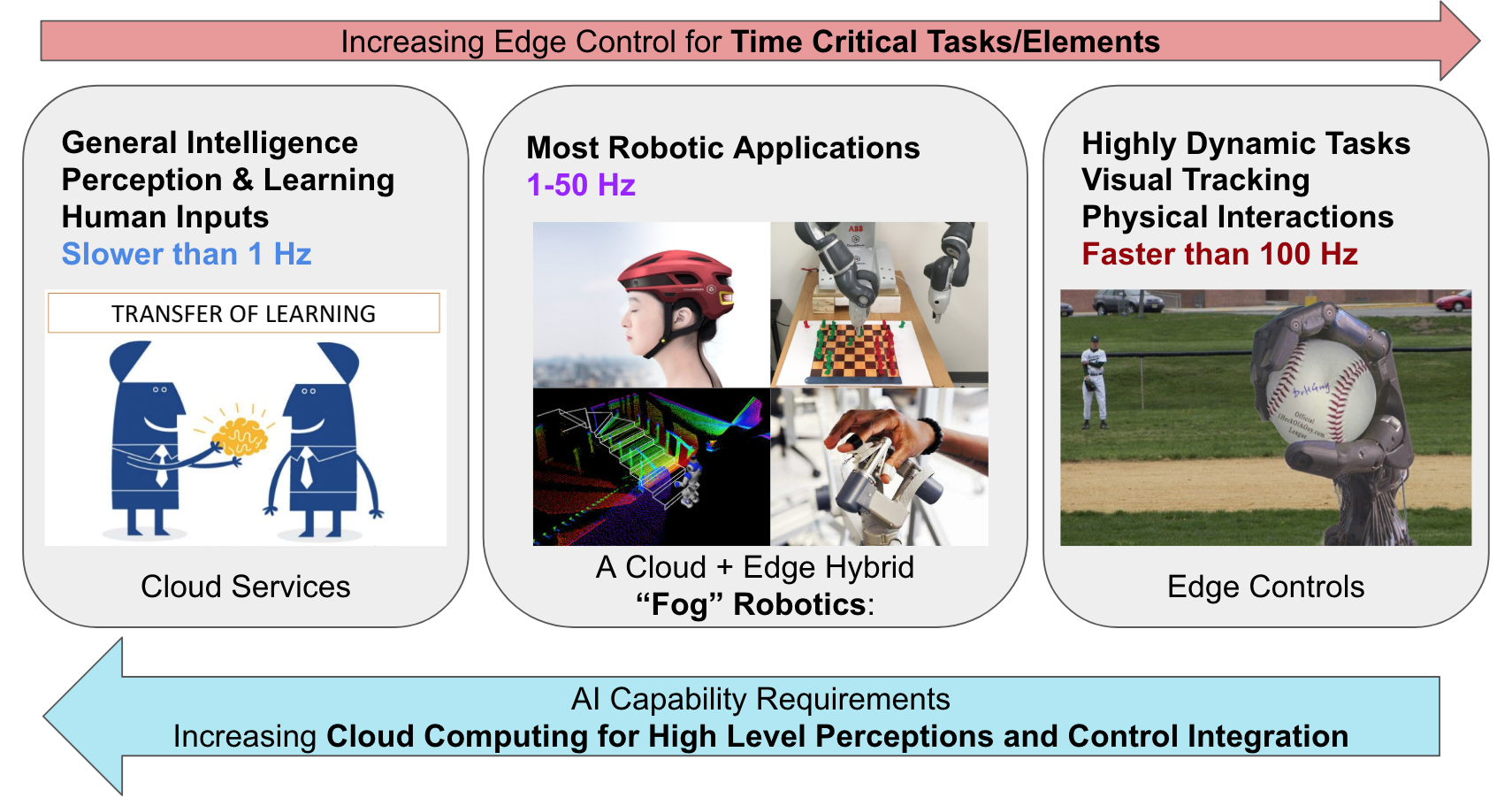}
	\caption { \textbf{Intelligence vs. Reflexes: Fog Robotics Balances Cloud and Edge Computation for Different Robotic Applications.}
		\textbf{(Left) Cloud Services } provide high-level intelligence, such as deep-learning based perception systems and cloud based human teleoperation.
		\textbf{(Right) Edge controller } fastest control loops, control highly dynamic robotic tasks, but often lacks high performance computer.
		\textbf{(Middle) Fog Robotics } combine the intelligence of the cloud and the responsiveness of the edge, covers 80\% of the robotic applications. 
		\textbf{(Top and Bottom)} \textbf{Move the computing to the cloud for more intelligence}, and \textbf{move the control to the edge for highly dynamic, closed-loop control}.  Together, Fog Robotic system can handle most service applications that require both intelligence and fast reflexes.
	}
	\label{fig:cloud_edge_app}
\end{figure*}

However, scalable cloud robotics systems are associated with high network communication costs, in the form of privacy, security, bandwidth, latency, and variability.  Specifically, network latencies and variabilities, bounded by speed-of-light and inconsistent network routings, prevent cloud-based robotic controller from controlling dynamic robots directly for interactive, human-compliant robot tasks, especially those requiring visual feedbacks.  

In this work, we combine both powerful cloud services and agile edge devices to build an intelligent Fog Robotic control system.  Our Fog Robotic system is implemented under Human Augmented Robotic Intelligence Platform, or HARI \cite{tian12cloud}, provided by Cloudminds Inc.  We integrate this system with a dynamic, dual arm, dual leg, self-balancing robot, named Igor, made by HEBI robotics.  

Leveraging a “heartbeat” communication protocol between cloud and edge, we are able to teleoperate Igor robustly using HARI.  We choose to perform a robotic task that is commonly performed in warehouse logistics, namely box pickups from a human carrier.  

While it is intuitive to teleoperate Igor during navigation, we found that it was extremely difficult and inefficient to teleoperate Igor to pick up objects as simple as a box.  To make the cloud-based teleoperation more intuitive, we program a dynamic automatic box-pickup module based on visual detection of an apriltag \cite{olson2011tags},\cite{wang2016iros}.  Apriltage detections are performed in the cloud to emulate how a cloud-based deep learning object recognition system would affect the performance of this Fog Robotic visual feedback loop.  Furthermore, to avoid performing time-consuming camera calibration and registration whenever the robot performs a box pickup, we choose to implement the module using 2D Image Based Visual Servoing (IBVS).  With the Fog Robotic IBVS, we demonstrate that Igor can perform reliable, automatic box pickups from a human carrier under unstructured environments.  

\section{CONTRIBUTION}

The main contributions of this work are as follows:
\begin{enumerate}
	\item \textbf{A ``heartbeat" protocol} that enables robust teleoperation of {a dynamic robot} in Fog Robotics.
	\item \textbf{A Fog Robotic visual servoing module} that enables automatic box-pickups to assist cloud teleoperators
	\item \textbf{Automatic box-pickups from a human} to demonstrate dynamic human robot interaction (HRI) under unstructured service environments.
\end{enumerate}

\section{RELATED WORK}


\textbf{Cloud Robotics} encompasses any robot or automation system that relies on either data or code from a network to support its operation \cite{kehoe15}. The term was introduced by James Kuffner in 2010.  It was evolved from \textit{Networked Robotics} ~\cite{kuffner10}. Well-known Cloud Robotic Systems includes: RoboEarth's Rapyuta \cite{rapyuta}, motion planning for services at both cloud \cite{vick15} and edge \cite{wafr2016alterovitz}, Berkeley robotics and automation as a service (Brass) \cite{tian2017cloud}, and Dex-Net as a Service (DNaaS) \cite{lidex}, just to name a few.  However, network costs in the form of privacy, security latency, bandwidth, and reliability present a challenge in Cloud Robotics \cite{Tanwani2018FogRoboticsDecluttering}.

\textbf{Fog Robotics} was recently introduced by Goldberg et al, and is defined as an extension of Cloud Robotics that balances storage, compute and networking resources between the Cloud and the
Edge \cite{Tanwani2018FogRoboticsDecluttering}.  It is inspired by Fog Computing, originally introduced by Cisco Systems in 2012 \cite{bonomi2012fog}.  In Fog Robotics, cloud computing resources is brought closer to the robot so that learning can be done close to where data is created. It has found its applications in service robots where robot can learn surface grasps from nearby unstructured environments ~\cite{Tanwani2018FogRoboticsDecluttering}.  In this work, we use Fog Robotics for (1) cloud-based teleoperation of a dynamic robot; (2) host vision servoing server to provide robot object recognition and localization feedbacks under unstructured environments.

\textbf{Robotic Vision} for service robots to operate under unstructured environment is challenging if traditional industrial robotic approaches were used, because these methods were developed for precision under highly structured manufacturing environments.  Registrations \cite{besl1992method} and calibration \cite{tsai1987versatile} are often done before a robotic task, and they can be time consuming and do require additional expertise for system maintaining.  

\textbf{Image Based Visual Servoings (IBVS)} \cite{hutchinson1996tutorial}, \cite{chaumette2006visual} uses camera 2D image space to measure relative distances between the robot and the target.  The measurements are independent of the exact 3D locations of the robot and the target.  Therefore, camera registrations and calibrations are not required before each robotic task, desirable for service robots deployments under unstructured environments for human robot interactions.  

Furthermore, recent development in deep-learning-based vision systems allow  recognition \cite{krizhevsky2012imagenet}, object detection \cite{girshick2015fast}, segmentation \cite{matterport_maskrcnn_2017}, and human gesture recognition \cite{cao2017realtime} to be performed for unstructured environments in semi-real-time (5 - 10Hz) on a GPU server.  Previously, we successfully implemented a cloud-based gesture perception system for a humanoid robot gesture mirroring task \cite{tian12cloud}.  More advanced robotic learning systems based on visual feedbacks has been developed for grasping \cite{dexnet16}, \cite{mahler2017dex}, visual servoing \cite{lee2017learning}, guided policy search \cite{levine2014learning}, \cite{levine2016learning}, visual foresight \cite{finn2017deep}, and domain randomization for transfer learning from simulation to reality \cite{tobin2017domain}, \cite{2018arXiv180800177O}.  All of these can be deployed in Fog Robotics to provide visual feed-backs for large scale service robot deployments.

\section{SELF-BALANCING ROBOT IGOR}

\begin{figure}[t]
	\centering
	\includegraphics[width=1.0\columnwidth]{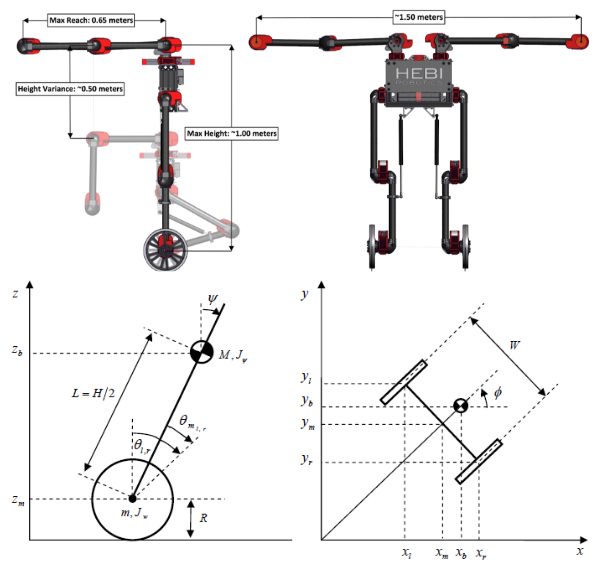}
	\caption{\textbf{Self-Balancing Robot Igor:} 
		\textbf{(Top)}14 Dof, dual-arm, dual-leg robot built with series elastic servo modules.
		 \textbf{(Bottom)} free body diagram of the inverted pendulum balancing control 
		 (\textbf{see supplemental video for details})
	}
	
	\label{fig:self_balancing_robot}
\end{figure}

We use a 14 degrees of freedom (DoF), dual-arm, dual-leg, dynamic self-balancing robot named Igor (shown in Fig. \ref{fig:self_balancing_robot}).  It is designed and made by HEBI robotics.  Each DoF is built with a self-contained, series elastic X-series servo modules.  These servos can be controlled with position, velocity, and torque commands simultaneously, and can provide accurate measurements of these three quantities to a central computer at high speed ( $>$1KHz) with minimum latency.  These modules are connected with Ethernet to an on-board Intel Nuc computer in the metal control box for self-balancing control.  

The self-balancing is achieved by modeling the system as an inverted pendulum (see Fig. \ref{fig:self_balancing_robot} bottom).  To estimate the center of mass (CoM) of the robot, the CoM of the two arms and two legs are first measured through HEBI's API in real-time using forward kinematics.  The position of the total CoM is then  estimated as the average of the CoMs of the four extremities plus the CoM of the control box weighted by the mass distribution:

\begin{equation} \label{center_of_mass}
x_{CoM} = \frac{{\sum_{i}{m{_i}x{_i}}}}{\sum_{i}{m{_i}}} \quad\quad i = \text{arms, legs, box}
\end{equation}

Igor also uses accelerometer measurements from the four servo modules attached to the control box to estimate the direction of gravity ($G$) at all times.  With CoM of Igor, center of wheels ($o_w$), and direction of gravity, we can calculate the length and direction of the inverted pendulum:
\begin{equation}
L = x_{CoM} - o_w
\end{equation}
\noindent The lean angle ($\psi$), which is the angle between gravity and the inverted pendulum can then be estimated in real-time:
\begin{equation}
\psi = cos^{-1}\Big(\frac{{L}\cdot{G}}{|{L}||{G}|}\Big)
\end{equation}

To keep the robot balancing, assuming that the lean angle ($\psi$) is small so that we can linearize the system:
\begin{equation}
\psi \approxeq sin(\psi)
\end{equation}

\noindent a torque ($\mathcal{T}$) in the direction of falling is applied to the wheel with radius ($R$) and angular velocity ($\omega$) to counteract the effects of gravity on the robot's center of mass:

\begin{equation}
\mathcal{T} = R\omega = v - \dot{\psi}{L}
\end{equation}

\noindent where $v$ is the velocity of the robot's CoM.  Furthermore, the derivative of the lean angle ($\dot{\psi}$) can be controlled by a proportional controller with coefficient ($K_v$), and is related to the velocity of the robot as follows:
\begin{equation}
\dot{\psi} = K_v{v}
\end{equation}

The real-time measurements of both robot CoM and direction of gravity are important, because the Igor controller needs to compensate for dynamic movements of the four extremities for robust self-balance control.  We can also rotate the robot by varying velocity applied to the two wheels, and control the angle of rotation based on inertia measurement unit (IMU) readings in real-time.

\section{AN INTELLIGENT FOG ROBOTIC CONTROLLER}

\begin{figure*}[th]
	\centering
	\includegraphics[width=0.7\textwidth]{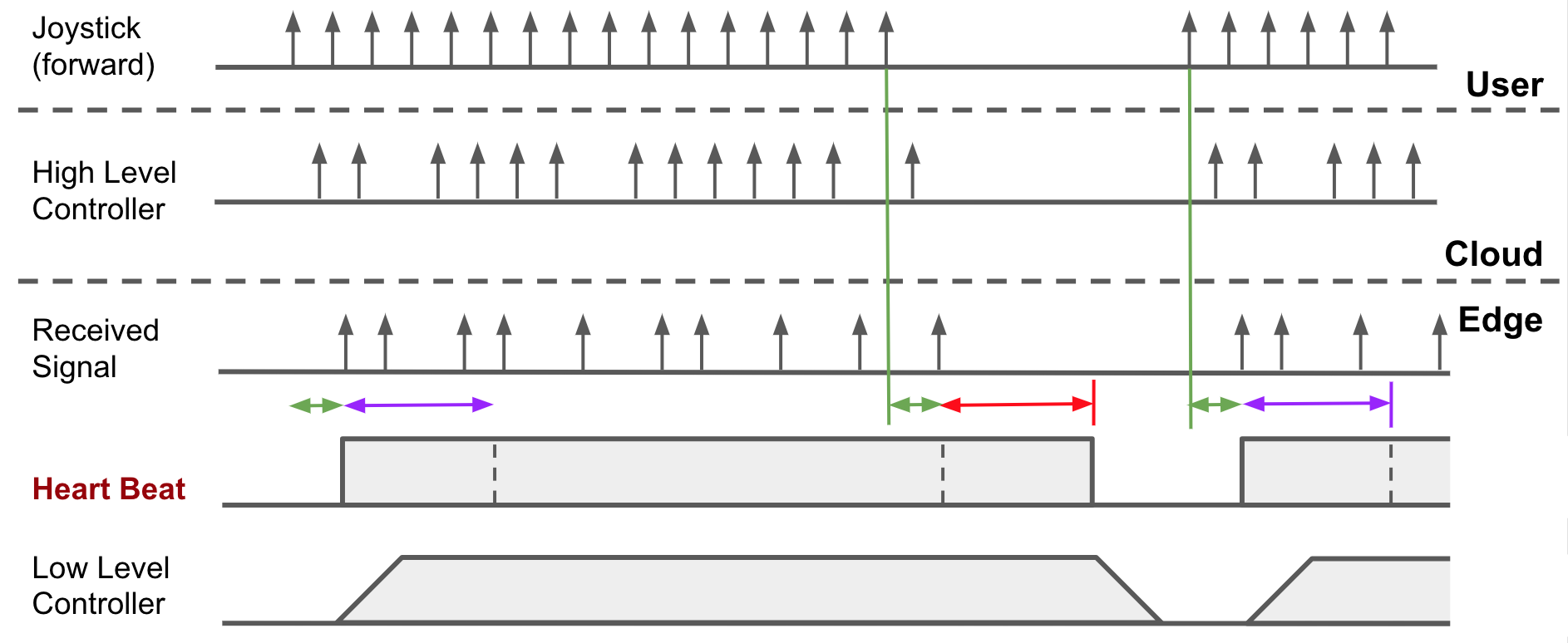}
	\caption{\textbf{``Heartbeat" Protocol with Asynchronous Communication to Send a Forward Command:}
		\textbf{(Top) Teleoperator with a Joystick} sending a series of forward command signals.
		\textbf{(Middle) Cloud High-Level Controller} receiving most of the packages with a little latency.  It then forwards received commands to the edge device.
		\textbf{(Bottom) Edge device} receives the commands with a lot of lost packages and some delays.  A ``heartbeat" signal for forward motion is turned on upon receiving the first command, and will stay on the duration of the sliding window \textbf{(250 ms, purple arrow)}.  The "heartbeat" will turn off when there is no command received during the sliding window \textbf{(red arrow)}.  The \textbf{green arrow} marks the network delay from the joystick to the edge device.  Finally, at the \textbf{low-level controller}, a ramp-up and ramp-down function is used to smooth out the start and stop forward velocity perturbation to the balancing system so that the robot can move forward without jitters or instability.
	}
	\label{fig:heartbeat}
\end{figure*}

\subsection{Edge Controllers}

There are two edge controllers in our system.  The first one is an Intel Nuc computer in the Igor control box.  It collects all sensor information from the 14 modular servos, and it controls all servos in real-time.  It hosts high-speed feedback control loops (200Hz or above) to maintain robot posture and self-balancing.  We refer to it as the low-level controller in Fig. \ref{fig:archetecture}.  

The other edge device is the robot command unit (RCU).  It is a smart android phone with a private LTE connection and a 2D camera (Fig. \ref{fig:archetecture}).  RCU serves as the gateway between the high-level cloud robotic platform and the low-level self-balancing controller.  It uses the private LTE connection to stream live videos to the cloud, and it receives and forwards high-level intelligent controls from the cloud to the low-level controller with minimum delay.  RCU works both indoors and outdoors with a good LTE reception.

Furthermore, we are in the process of integrating HEBI's self-balancing controller into the smart phone.  It can replace the native Intel Nuc computer so that it will serve as both RCU and low-level controller.  By making the edge controller more compact, we gain more battery room in the control box for a longer robot operation time.

\subsection{Cloud Controller}

A high-level intelligent robot controller is placed in the cloud to work with the edge controller (see Fig. \ref{fig:archetecture}).  It operates at a lower speed (3-5Hz), yet it commands the robot based on HARI's Artificial Intelligence (AI) and Human Intelligence (HI) services, which is critical for robots to operate under unstructured environments.  Depending on the situation, it can either extract commands based on the object recognition server or forward commands sent from a cloud teleoperator.  These high-level commands are sent to RCU and are executed in a different form on the low-level controller.

\subsection{Hybrid Control with ``Heartbeat"}

When controlling Igor, commands sent from the high-level cloud controller act as perturbations to a time-invariant, stable system maintained by the low-level self-balancing controller at the edge.  This is a form of hybrid control where discrete signals are sent from the cloud to control a dynamical system at the edge.

Minimum delays in the cloud to edge robot command delivery are desirable for intuitive teleoperation.  We choose to use the asynchronous network protocol UDP to implement the cloud-edge communication.  However, since deliveries are not guaranteed in UDP, packages can be lost during communication.  Further, the packages can arrive at the designation in different orders from their original sequence.  Both problems can create variable controls at the edge controller, which can cause instability in self-balancing.  This would affect user experiences during teleoperation as well, and can be dangerous to people around the robot.  

We implement a ``heartbeat" signal at the edge controller to solve these problems (shown in Fig. \ref{fig:heartbeat}).  The ``heartbeat" is a switch signal that is turned on when the first signal arrives at the edge.  It will remain on for a period of time (t) and will only turn off if there is no package received for the selected command during this time.  We can view the ``heartbeat" design as performing a ``convolution" with a moving window on the signal received. Finally, we turn the ``heartbeat" signal into an edge control signal with a ramping function at the beginning and the end of the control to ensure a smooth start and end action when the controlled perturbation hits the stable self-balancing system.  

\section{DYNAMIC VISUAL SERVOING}

To assist teleoperation with automation, we focus on using Fog Robotics to control a dynamic robot with Image Based Visual Servoing (IBVS) to automatically pick up a box. We choose IBVS because it eliminates extensive camera calibration that is hard to maintain on a dynamic robotic system.  The goal of our IBVS is to navigate the robot to an optimal box pickup location where the apriltag lay within the green target box, which has the same size as the apriltag (Fig. \ref{fig:visual_servoing})

The aim of visual-servoing-based control is to minimize the relative error between the measured target position and desired target position $\boldsymbol{e}(t)$: 
\begin{equation} \label{visual_servo}
\boldsymbol{e}(t) = \boldsymbol{s}(\boldsymbol{m}(t),\boldsymbol{a}) - \boldsymbol{s}^*
\end{equation}

\noindent where $\boldsymbol{m}(t)$ is a set of image measurements and $\boldsymbol{a}$ is a set of parameters, such as camera intrinsics, that represents additional knowledge about the system.  $\boldsymbol{s}$ is the measured values of image features/object locations, such as pixel coordinates in the picture frame, and $\boldsymbol{s}^*$ is the desired values of image features/object locations.

The change of feature error $\boldsymbol{\dot{e}}$ and camera velocity $\boldsymbol{v_c}$ is related by \textit{interactive matrix} $\boldsymbol{L}$:

\begin{equation}
\boldsymbol{\dot{e}} = \boldsymbol{L}\boldsymbol{v_c}
\end{equation}

For IBVS, which is done in 2D image space, 3D points $\boldsymbol{X}=(X, Y, Z)$ are projected onto 2-D images with coordinates $\boldsymbol{x}=(x, y)$:
\begin{align} 
x = X/Z\\
y = Y/Z 
\end{align}

\noindent which creates an interactive matrix for 2D image based servoing:
\[
\boldsymbol{L} = 
\begin{bmatrix}
 -1/Z & 0 		& x/Z 	& xy 	& -(1+x^2) 	&y\\
 0    & -1/Z 	& y/Z	& 1+y^2 & -xy		&-x
\end{bmatrix}
\]


\noindent With the interactive matrix, camera velocity can be estimated by:
\begin{equation}
\boldsymbol{v_c} = -\lambda{\boldsymbol{L}^{+}}\boldsymbol{e} = -\lambda{\boldsymbol{L}^{+}}(\boldsymbol{s} - \boldsymbol{s}^*)
\end{equation}

\noindent where $L^{+}$ is the Moore-Penrose pseudo-inverse of $L$:
\begin{equation}
\boldsymbol{L}^+ = (\boldsymbol{L}^T\boldsymbol{L})^{-1}\boldsymbol{L}^T
\end{equation}

The final control law is set as a robot velocity effort $v_s$ opposite to the camera velocity $v_c$ because the target moves in the opposite direction of the camera in the image frame:
\begin{equation}
\boldsymbol{v_s} = -\boldsymbol{v_c}
\end{equation}

\noindent Notice that the interactive matrix depends only on $x$ and $y$, that is the 2D pixel coordinate of the target, and $Z$ which is the depth of the target.  In our system, $Z$ is measured as the size of the apriltag.  Therefore, the IBVS measurement is independent of the exact 3D position of the target measurement, which is attractive to our system because the exact 3D camera registration is not required.

\subsection{IBVS Implementations for Automatic Box Pickup}

\begin{figure}[t]
\centering
\includegraphics[width=1.0\columnwidth]{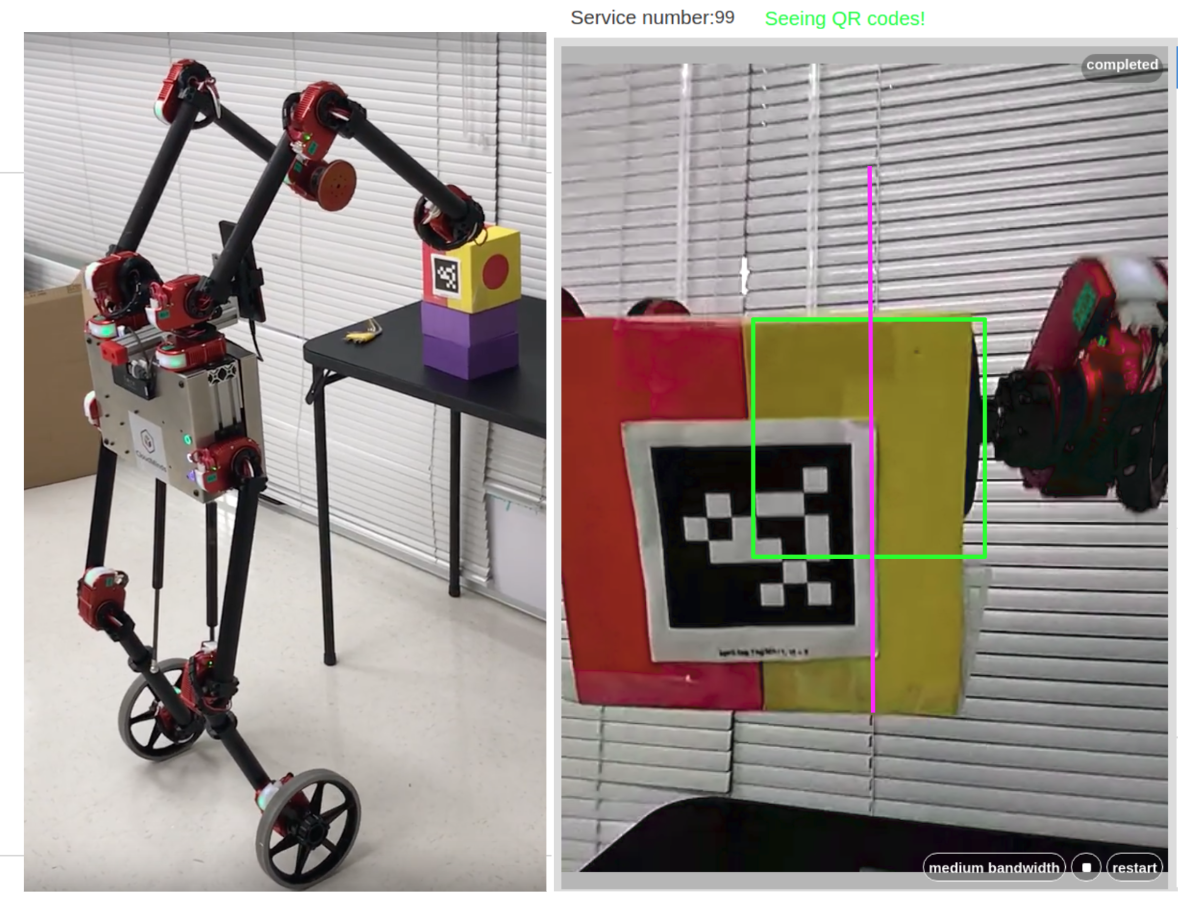}
\caption{\textbf{Image Based Visual Servoing using Fog Robotics: }
                \textbf{(Left) Igor Picking Up a Static Box;}
                \textbf{(Right) Image Based Visual Servoing (IBVS)} the purple central line and the size of the green box are used in the phase 1 of the IBVS.  Phase 1 controls how to navigate the robot an optimal position.  Phase 2 controls the optimal height of the robot.  The robot changes its height by the two ``knee" joints of the two legs so that the apriltag would fit directly into the green box in the center of the 2D image frame.  \quad
                \textbf{Videos: (1) Pickup a box from a table: \url{https://youtu.be/b0mr5GHHjBg} (2) Pickup a box from a human: \url{https://youtu.be/K9R4y2w1uPw}}
}
\label{fig:visual_servoing}
\end{figure}

The automatic visual servoing controller executes a box pickup in three phases.  Phase 1 uses IBVS to move the robot to a position where the aprialtag has the same size as the green box shown in left side of Fig \ref{fig:visual_servoing}.  The robot also need to position apriltag on the center purple line of the video frame after phase one, but not at the center.  In phase 2, the robot adjusts its own height by changing the joint angles of the two ``knee" joints so that the apriltag would lay at the center of the video frame where the green box is.  After the robot reaches the optimal picking position when apriltag is at the center of the video, phase 3 begins.  The robot controller commits to perform a box-pickup with a pre-defined, hard-coded, dual arm, grasping motion.  

The size of the target green box encodes depth information ($Z$) of the target.  We find this optimal size by teleoperating the robot to different positions that is close to the box.  From locations, we select the position that has the highest box-pickup successful rate using the hard-coded grasping motion.  

\subsection{Fog Robotic IBVS}

Although we use simple apriltags for object recognition, we aim to anticipate the design of a deep-learning-based Fog Robotic visual system for robotic pickups. Therefore, to emulate the latency effects under such system, we deploy apriltag recognition in the cloud and use ``heartbeat” protocol to stream apriltag’s
geometric locations to low-level controller via RCU. Together, we build a robust Fog Robotic IBVS controller for box pickups.

\section{EXPERIMENTS AND RESULTS}

With the ``heartbeat" design, we are able to navigate the self-balancing robot reliably (see video) from the cloud-based teleoperation interface.  We also pre-program arm actions such as clapping, and ``disco" for human robot communications and entertainment (video: \url{https://youtu.be/1H1VEpkbG_E}).  

We further hardcode a box pickup motion for the two arms, and attempt to pick up a box via the cloud-based teleoperation.  However, even with a reliable teleoperation module and with pre-programmed pick up motions, we find it extremely difficult and inefficient to pick up a box using cloud teleoperation.  We suspect that it is caused by the lack of natural, immersive 3D visual perception for the teleoperator,  

To quantify the observation, we perform two different teleoperation experiments with 10 trials each: 1) control Igor locally so that the operator can see the robot and the box; 2) control Igor from the cloud to pick up the box.  In both cases, the box is positioned on the table in a stable location.  The robot is about 2 meters from the object and faces the front of the box (see video \url{https://youtu.be/b0mr5GHHjBg}).  We observe that local teleoperator (at least 2 meters away from the target) can perform box pickups much faster with a higher success rate than the cloud teleoperator (see table \ref{tab:recogn_accuracy})

After implementing the automatic IBVS module, we perform the same experiments and benchmark the automatic module with human in the loop.  We allow the cloud operator to first teleoperate the robot as fast as possible to a location where apriltag is recognizable, which can be up to  20 degrees from the surface normal direction of the apriltag.  Then, the human triggers the Fog Robotic IBVS module, allowing the robot to pick up the box automatically.  We observe that the speed of this third case is on-par with human local teleoperation, but the reliability is even higher at 100\% (see table  \ref{tab:recogn_accuracy})

\begin{table}[h]
	\caption{Teleoperation vs. Automatic Box Pickups}
	\centering
	\begin{tabular}{ p{1.8cm} S[table-format=2] S[table-format=2]  }
		\hline 
		& {Average Duration (s)} & {Success Rate } \\
		\hline\hline  
		Local Teleop & 43 & 9/10  \\
		Cloud Teleop & 340 & 4/10 \\
		\hline
		Auto Pickups  & 46 & 10/10 \\
		\hline
	\end{tabular}
	\label{tab:recogn_accuracy}
\end{table}

Finally, we leverage the flexibility of visual servoing to perform a proximate human robot interaction (HRI) task.  During the task, the human carries the box with an apriltag.  They can show Igor the apriltag while moving around.  Igor can recognize and localize the tag with a distance as far as 6 meters.  As soon as the robot  recognizes the apriltag, the teleoperator can release the robot so that it enters automatic mode.  We observe that as long as the robot can recognize the apriltag and the box is in a reachable height for the robot, the robot will follow the human around, and eventually pick up the box from the person with a high successful rate (see video: \url{https://youtu.be/K9R4y2w1uPw})

\section{DISCUSSION AND FUTURE WORK} 

In this work, we take advantage of both intelligent cloud robotic platform and edge controller to build an intelligent Fog Robotic system that can perform human-compliant, automatic box pickups using visual servoing.  

A ``heartbeat" protocol with asynchronous communication is introduced to mitigate network latencies and variabilities effects on the dynamic hybrid self-balancing controller in Fog Robotics.  However, the current ``heartbeat" protocol is not perfect.  There is an increased delay at the end of command signal that would cause a delayed reaction after the last command signal is received (see red arrow in Fig. \ref{fig:heartbeat}). This imposes a significant safety concern, because even if the ``heartbeat" time window is short, i.e. 250 ms in our case, the robot will not stop completely until after 250 ms plus the ramping down period.  To compensate, we implement a sharper ramp function at the stop compared to the ramp function at the start, but 250 ms is the hard limit for the delay on the current system (shown as red arrow in Fig. \ref{fig:heartbeat}).  

Our future work includes a better modeling of package drops in asynchronous communication, so that we can build a probability model to measure variabilities of time intervals between packages.  This way, we can further reduce this delayed reaction by adjusting the ``heartbeat" window size based on the predicted time of last package.

Like other service robots, Igor needs to interact and cooperate with human beings.  We demonstrate the advantages of visual servoing: (1) it requires no calibrations before each robotic task; (2) it can handle dynamic human robot interaction, such as following a human to pick up a box from that person.  Our automatic system works even in  unstructured environments when obstacles are present between the human and the robot. The self-balancing control and the compliant servos can correct themselves when small obstacles are encountered.  The human box carrier or the cloud teleoperator can also help the robot avoid obstacles by guiding it to a path with more clearance. 

One failure case is when the human carrier tricks the robot.  It happens if the target is moved after the robot commits to the final phase of box picking, which is hard-coded.  


In the future, we can program a more dynamic automatic object pickup so that the robot can pick up a moving object with a continuous motion, without hard-codings.  We also plan to deploy deep-learning recognition pipelines such as mask-RCNN (cite) together with intelligent grasping systems such as dex-net \cite{dexnet16}\cite{mahler2017dex}\cite{lidex} using Fog Robotic systems, so that it can guide both dynamic robots such as Igor and static robots such as YuMi\cite{tian2017cloud} and HSR \cite{laskey2017dart} to perform generalized, human compliant object pickups and manipulations.

\section{ACKNOWLEDGMENTS}

We thank members from the AUTOLAB-- Ron Berenstein, Ajay Tanwani--for discussions, and members at Cloudminds--Arvin Zhang, Havelet Zhang, Mel Zhao--for their technical support.  Special thanks Prof. Joseph Gonzalez for discussions on hybrid synchronous and asynchronous Systems.  We thank Matthew Tesch, David Rollinson, Curtis Layton, and Prof. Howie Choset from HEBI robotics for continuous support, training, and discussion on the Igor self-balancing research robot.

Fundings from Office of Naval Research, NSF EPCN, and Cloudminds Inc. are acknowledged.  Any opinions, findings, and conclusions or recommendations expressed in this material are those of the
authors and do not necessarily reflect the views of the Sponsors.

\bibliographystyle{IEEEtran}

\newpage

\bibliography{references}

\end{document}